\documentclass[10pt,twocolumn,letterpaper]{article}

\usepackage{cvpr}
\usepackage{times}
\usepackage{epsfig}
\usepackage{graphicx}
\usepackage{amsmath}
\usepackage{amssymb}
\usepackage{subfigure}
\usepackage{multirow}
\usepackage{threeparttable}
\usepackage{diagbox}
\usepackage{color}
\usepackage{authblk}

\usepackage[pagebackref=true,breaklinks=true,letterpaper=true,colorlinks,bookmarks=false]{hyperref}

\cvprfinalcopy 

\def\cvprPaperID{4702} 

\ifcvprfinal\fi
\begin{document}

\title{COCAS: A Large-Scale Clothes Changing Person Dataset for Re-identification}

\author[1,2]{Shijie Yu$^*$}
\author[3]{Shihua Li\thanks{Equally-contributed first authors(sj.yu@siat.ac.cn,lishihua@ime.ac.cn)}}
\author[1]{Dapeng Chen}
\author[1]{Rui Zhao}
\author[1]{Junjie Yan}
\author[1]{Yu Qiao\thanks{Corresponding author (yu.qiao@siat.ac.cn)}}
\affil[1]{ShenZhen Key Lab of Computer Vision and Pattern Recognition, SIAT-SenseTime Joint Lab,Shenzhen Institutes of Advanced Technology, Chinese Academy of Science}
\affil[2]{University of Chinese Academy of Sciences, China}
\affil[3]{Institute of Microelectronics of the Chinese Academy of Sciences}

\maketitle
\thispagestyle{empty}

\begin{abstract}
  Recent years have witnessed great progress in person re-identification (re-id). Several academic benchmarks such as Market1501, CUHK03 and DukeMTMC play important roles to promote the re-id research. To our best knowledge, all the existing benchmarks assume the same person will have the same clothes.  While in real-world scenarios, it is very often for a person to change clothes. To address the clothes changing person re-id problem, we construct a novel large-scale re-id benchmark named \textbf{C}l\textbf{O}thes \textbf{C}h\textbf{A}nging Person \textbf{S}et (COCAS), which provides multiple images of the same identity with different clothes. COCAS totally contains 62,382 body images from 5,266 persons. Based on COCAS, we introduce a new person re-id setting for clothes changing problem, where the query includes both a clothes template and a person image taking another clothes. Moreover, we propose a two-branch network named  Biometric-Clothes Network (BC-Net) which can effectively integrate biometric and clothes feature for re-id under our setting. Experiments show that it is feasible for clothes changing re-id with clothes templates.
\end{abstract}

\section{Introduction}

\emph{``On Tuesday, December 29, 2015, a white female suspect walked into the Comerica Bank, stating she was armed with a bomb and demanded money. The female suspect escaped with an undisclosed amount of cash. The video shows that the suspect run behind the laundromat, change clothes and flee north towards the I-94 Service Drive. ''}\footnote{http://www.wjr.com/2016/01/06/woman-wanted-in-southwest-detroit-bank-robbery}

\vspace{0.5em}

\begin{figure}[htbp]
    \centering
    \subfigure[A realistic scenario for clothes changing person re-id.]{
        \begin{minipage}[t]{0.48\textwidth}
            \centering
            \includegraphics[width=1.0\textwidth]{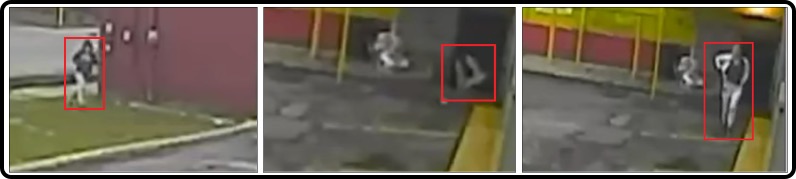}
        \end{minipage}
    }
    \subfigure[Example of our clothes changing re-id setting]{
        \begin{minipage}[t]{0.48\textwidth}
            \centering
            \includegraphics[width=1.0\textwidth]{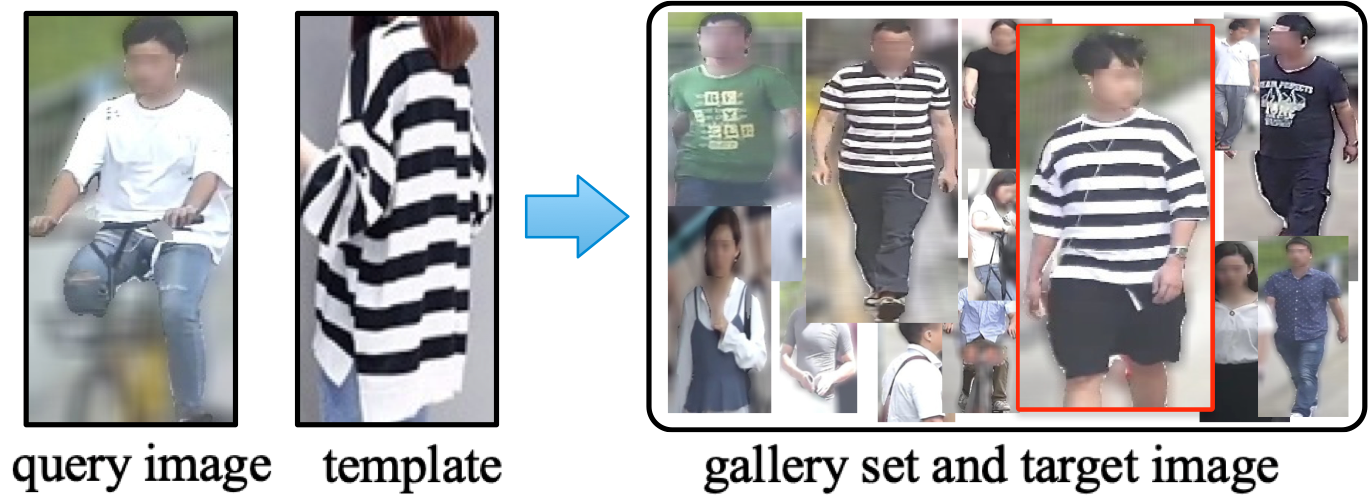}
        \end{minipage}
    }
    \centering
    \caption{(a) shows a realistic case that a suspect with a black coat changed her clothes to a white coat. (b) shows our clothes changing re-id setting,  where the man with white and black stripes T-shirt (the target image which is marked by a red box) is identified by the man with white T-shirt (the query image) and the white and black stripes T-shirt (the clothes template).} 
    \vspace{-0.5cm}
    \label{fig:teaser}
 \end{figure}
 
 \makeatletter 
  \newcommand\figcaption{\def\@captype{figure}\caption} 
  \newcommand\tabcaption{\def\@captype{table}\caption} 
\makeatother
\begin{figure*}[t]
\centering
    \begin{minipage}{0.58\textwidth}
    \centering
        \includegraphics[width=0.96\textwidth]{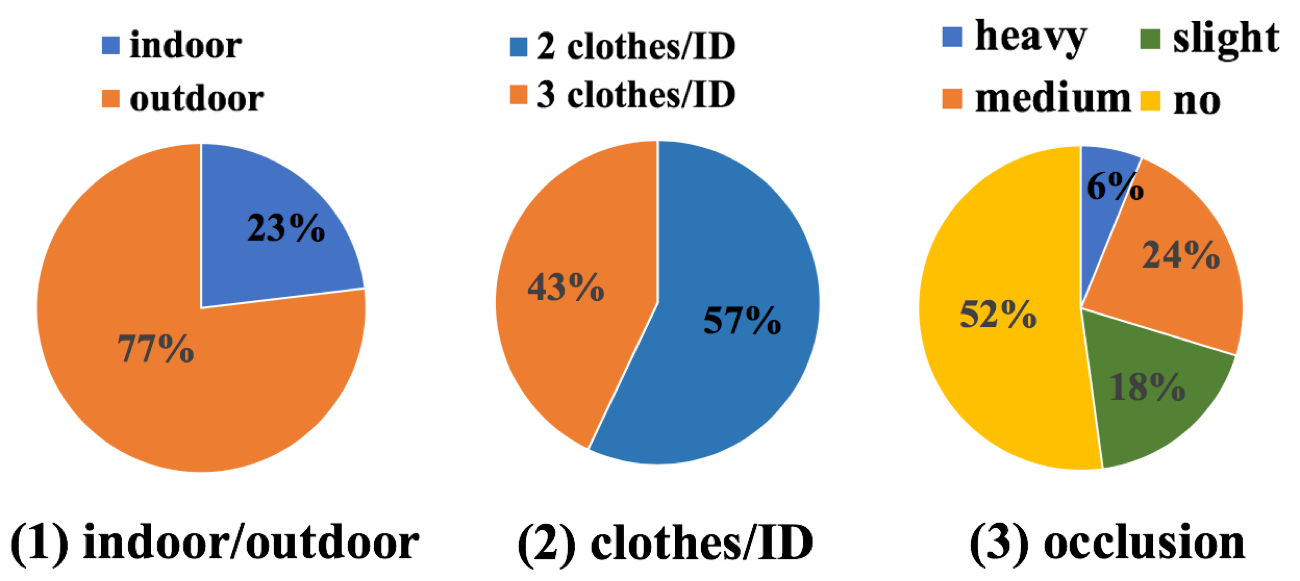}
        \vspace{0.8em}
        \figcaption{Statistical information of COCAS.  }
        \label{fig:statistics}
    \end{minipage}
\hspace{0.5em}
\begin{minipage}{0.38\textwidth}  
    \centering
    \includegraphics[width=0.98\textwidth]{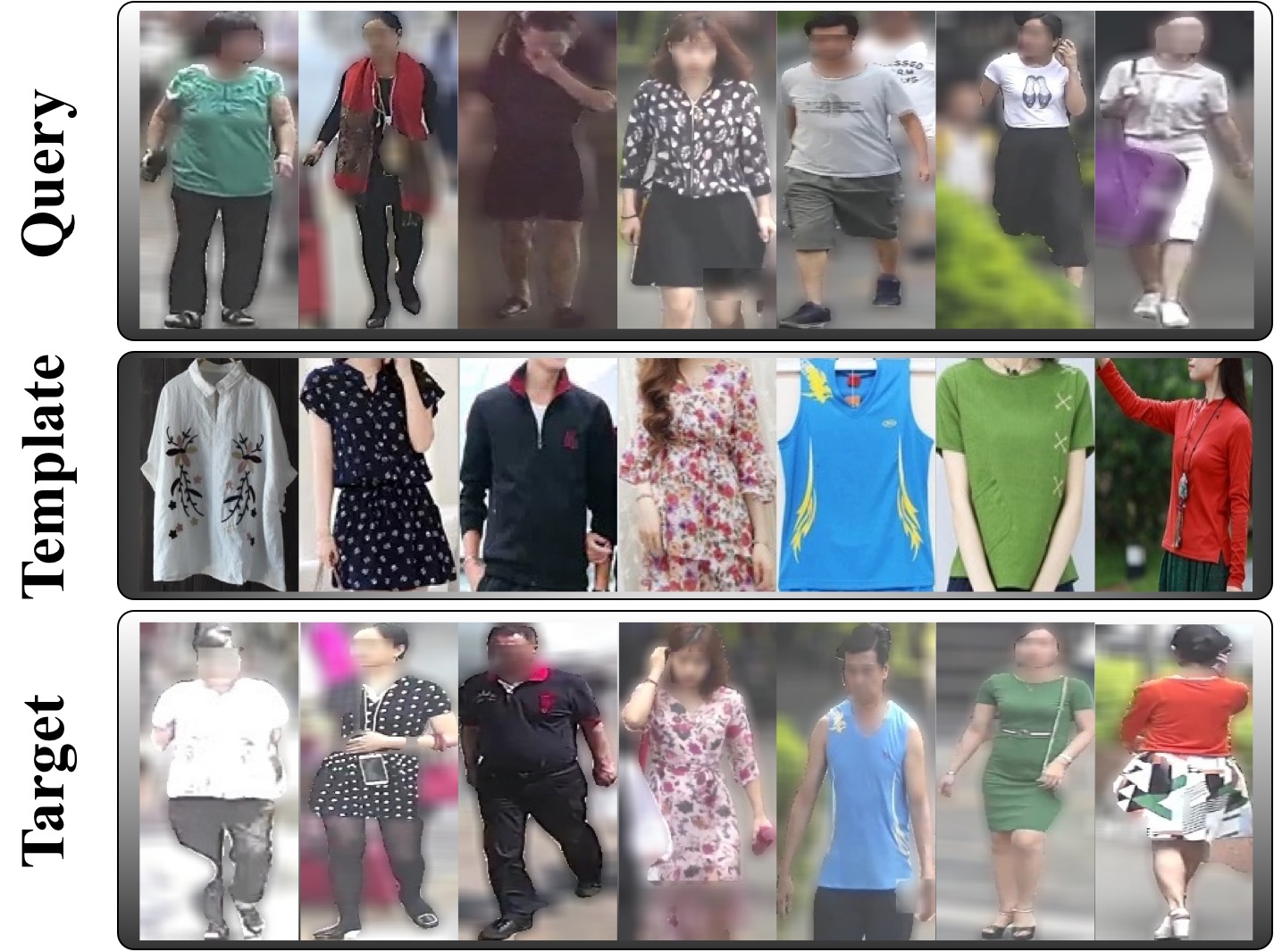}
     \figcaption{Instances in COCAS dataset. }
    \label{fig:dataset}
\end{minipage}
\end{figure*}
\begin{table*}[htbp]
    \centering
    \caption{Comparing COCAS with public re-id datasets}
    \label{tab:comparison}
    \scalebox{0.85}{
    \begin{tabular}{l|c|c|c|c|c|c|c|c}
        \hline
        \textbf{Dataset}&\textbf{VIPeR\cite{Gray2007EvaluatingAM}}&\textbf{ETHZ\cite{schwartz09d}}&\textbf{CUHK01\cite{li2012human}}&\textbf{CUHK03\cite{li2014deepreid}}&\textbf{Market 1501\cite{zheng2015scalable}}&\textbf{Airport\cite{karanam2016Airport}}&\textbf{DukeMTMC\cite{Zheng_2017_ICCV_Duke}}&\textbf{COCAS}\\
        \hline\hline
        \textbf{ID num}&632&148&971&1,467&1,501&9,651&1,812&5,266\\\hline
        \textbf{BBox num}&1,264&8,580&3,884&14,096&32,668&39,902&36,411&62,382\\\hline
        \textbf{Body img}&hand&hand&hand&hand&DPM&ACF&hand&hand\\\hline
        \textbf{Multi-Clot.}&No&No&No&No&No&No&No&Yes\\
        \hline
    \end{tabular}
    }
    \vspace{-0.5cm}
\end{table*}

Person re-identification (re-id) plays more and more important roles in video surveillance systems with a wide range of applications. Previous re-id protocols \cite{zheng2015scalable, zheng2016Mars, li2014deepreid, li2013locally, li2012human, Wei_2018_CVPR, Chen_2016_CVPR, Chen_2015_CVPR, Chen_2020_ICLR, Chen_2019_ICCV} assume the same person wears the same clothes, however, they cannot cope with the clothes changing case in the above news. 
As shown in Fig. \ref{fig:teaser}a, the suspect wants to escape being arrested,  thus she intentionally changes the clothes. The conventional re-id models \cite{Chen_2018_CVPR, Sun_2018_ECCV, Chen_2019_ICCV, zheng2015scalable, He_2018_CVPR_Partial, Si_2018_CVPR_reid, Xu_2018_CVPR_reid} tend to fail for at least two reasons. First, these models are trained on identities with the same clothes. The clothes' appearance is statistically regarded as a kind of discriminative feature by the model. Furthermore, biometric traits like face and body shape are too weak to be learned, because they only take up a small part of the body image. On the other hand, learning a clothes-irrelevant person re-id model is also difficult. The model can hardly be applied over the large-scale person image set if we only utilize unclear face and body shape information.

We instead consider an easier but practical clothes changing re-id problem as shown in Fig. \ref{fig:teaser}b: a person image (target image) is searched by a clothes template image and an image of the same person wearing another clothes (query image). Compared with the current re-id setting, such setting makes use of clothes templates and still has a wide range of real applications. Take two examples, for finding the lost children/elders, the family just needs to provide a recent photo of the lost child/elder and an image of clothes the child/elder wears also. For clothes changed suspect tracking, the police can find more relevant images by a captured image and a clothes template described by a witness or left by the suspect.

To tackle the problem, we build a large-scale benchmark, named \textbf{C}l\textbf{O}thes \textbf{C}h\textbf{A}nging Person \textbf{S}et (COCAS). The benchmark contains 62,832 body images of 5,266 persons, and each person has 5$\thicksim$25 images with 2$\thicksim$3 clothes. For each person, we move the images of one kind of clothes into the gallery set, and these images are target images.  We further find a clothes template from the internet, which is similar to the clothes in these target images of that person. All the remaining images are put into the 
query set. In fact, collecting such a large-scale person-related dataset is non-trivial. We will detailedly describe data collection, person and clothes association,  privacy protection strategies and protocol definition in section \ref{COCAS benchmark}. Biometric-Clothes Network (BC-Net) with two branches is also proposed to address clothes changing re-id. 
One branch extracts biometric characteristics of a person such as the face, body shape and hairstyle. The other branch extracts clothes feature,  whose inputs are different for the query image and target image. A query image utilizes the clothes template for the branch to better match the target image, while a target image employs the clothes detection module to obtain a clothes image patch from itself. 

In summary, our contributions are three-folder: (1) We define a kind of clothes changing re-id problem where the queries are composed of person images and clothes templates. (2) A novel large-scale dataset named COCAS is built for clothes changing re-id. (3) We propose BC-Net that can separate the clothes-relevant and clothes-irrelevant information, making the changing clothes re-id problem feasible by providing clothes templates of target images. Interesting ablation studies are conducted including examining how clothes appearance can influence the re-id.  The performance of BC-Net indicates the clothes changing re-id is promising by employing the clothes templates.

\begin{figure*}[t]
    \centering
    \includegraphics[width=1.0\textwidth]{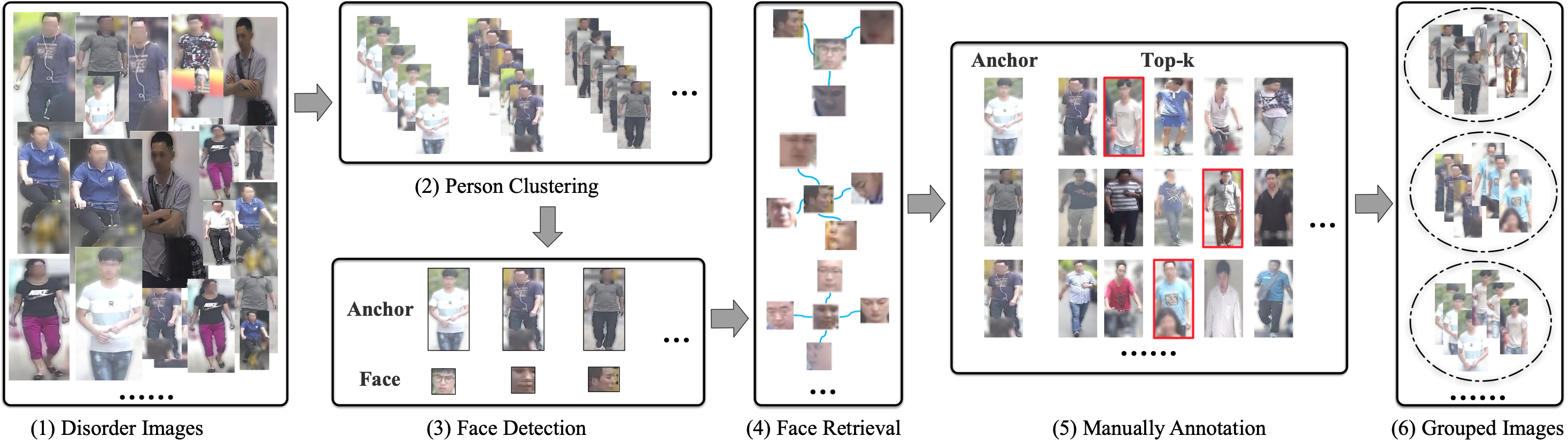}
    \caption{The pipeline of associating images of the same person. (1) The dataset with disordered body images. (2) Person image clustering. (3) Face detection and anchor image selection. (4) $K$-nearest neighbours searching by facial feature. (5) Manually annotation on the $K$-nearest neighbours to choose the truly matched person. The red boxes are examples to show the annotation results. (6) The final dataset, where each person has different types of clothes. }
    \label{fig:process}
    \vspace{-0.5em}
\end{figure*}


\section{Related Work}
\noindent\textbf{Re-ID Datasets.} Most recent studies on person re-id are based upon data-driven methods \cite{zheng2015scalable, He_2018_CVPR_Partial, Zheng_2015_ICCV_Partial, Si_2018_CVPR_reid, Xu_2018_CVPR_reid}, and there emerges a lot of person re-identification datasets. Among these datasets, VIPeR \cite{Gray2007EvaluatingAM} is the earliest and most common dataset, containing 632 identities and 1,264 images captured by 2 cameras.  ETHZ \cite{schwartz09d} and  RAiD \cite{Das2014} contains 8,580 images of 148 identities and 6,920 images of 43 identities, respectively. One insufficiency of these datasets is that the data scale is too small so that they cannot fully support the training of deep neural network. Several large-scale datasets, including CUHK03 \cite{li2014deepreid}, Market1501 \cite{zheng2015scalable} and DukeMTMC \cite{Zheng_2017_ICCV_Duke}, \textit{etc.}, become popular along with the development of the deep neural networks. As the performance gain is gradually saturated on the above datasets, newly proposed datasets become larger and larger including Airport \cite{karanam2016Airport}, MSMT17 \cite{Wei_2018_CVPR} and RPIfield \cite{Zheng_2018_CVPR_Workshops}. 

\noindent\textbf{Re-ID Methods.}
Early works on person re-id concentrated on either feature extraction \cite{wang2007shape, ma2012bicov, corvee2010person, farenzena2010person} or metric learning \cite{koestinger2012large, li2013learning, chen2016similarity, mignon2012pcca}. Recent methods mainly benefits from the advances of CNN architectures, which learn the two aspects in an end-to-end fashion \cite{Chen_2017_CVPR_deep,Li_2018_CVPR_deep, Li_2017_CVPR_deep, Wang_2016_CVPR_deep, Xu_2018_CVPR_reid, Xiao_2016_CVPR_deep}. Our work can be positioned as a kind of deep neural network. Typically, the re-id problem can be simply trained with the identification loss, by treating each person as a specific class \cite{Xiao_2017_CVPR}. With the deep similarity learning, person re-id is to train a siamese network with contrastive loss \cite{varior2016gated, Varior2016}, where the task is to reduce the distances between images of the same person and to enlarge the distances between the images of different persons.  Several methods employed triplet loss to enforce the correct order of relative distances among image triplets, \emph{i.e.}, the positive image pair is more similar than the negative image pair w.r.t. a same anchor image \cite{Cheng_2016_CVPR_reid, Chang_2018_CVPR_reid}.  Our method is different from the previous re-id methods, where the query includes both person image and clothes template. A two-branch network is employed to extract the biometric feature and clothes feature, supervised by both identification loss and triplet loss.  

\noindent\textbf{Clothes Changing in Re-ID.}
Clothes changing is an inevitable topic when it comes to re-id, but there are still less studies about it due to lack of large-scale realistic datasets. There are several related works on clothes changing re-id. Xue \textit{et el.} \cite{Xue2018ClothingCA} focused on re-id in photos based on the People in Photo Album (PIPA) dataset \cite{Zhang_2015_CVPR_PIPA}. The work in \cite{LiDangwei} is based on a sub-dataset built from RAP \cite{LiDangwei}, where 235 identities have changed clothes. Furthermore, Generative Adversial Network (GAN) is also applied to clothes changing. Jetchev \textit{et el.} \cite{Jetchev_2017_ICCV} introduced conditional analogy GAN to changed the person's clothes to target clothes. Zheng \textit{et el.} \cite{Zheng_2019_CVPR} proposed to disentangle the person images into appearance and structure, then reassembled them to generate new person images with different clothes.



\section{The COCAS Benchmark} \label{COCAS benchmark}

COCAS is a large-scale person re-id benchmark that has different clothes for the same person.  It contains 5,266 identities, and each identity has an average of 12 images. The images are captured in diverse realistic scenarios (30 cameras), including different illumination conditions (indoor and outdoor) and different occlusion conditions. The comparison with the existing re-id benchmark is shown in Tab. \ref{tab:comparison} and several instances fetched from COCAS dataset are shown in Fig. \ref{fig:dataset}.

 \noindent \textbf{Data Collection.} COCAS was collected in several commodity trading markets where we got permission to place 30 cameras indoors and outdoors. We recruited the people who did not mind being presented in the dataset (we promised to blur the facial region for personal privacy). As there was a high flow of people, a sufficient number of identities can be observed. As some people came to the markets every day and the data was collected in 4 different days, there were great chances to capture their images with different clothes. Clothes template images were acquired based on the collected person images. We first cropped the clothes patches from the person images by the a human parsing model, LIP \cite{Gong_2017_CVPR}, and searched the corresponding clothes templates by the image search engine from the Internet.

\begin{figure}[t]
    \centering
    \includegraphics[width=1.0\linewidth]{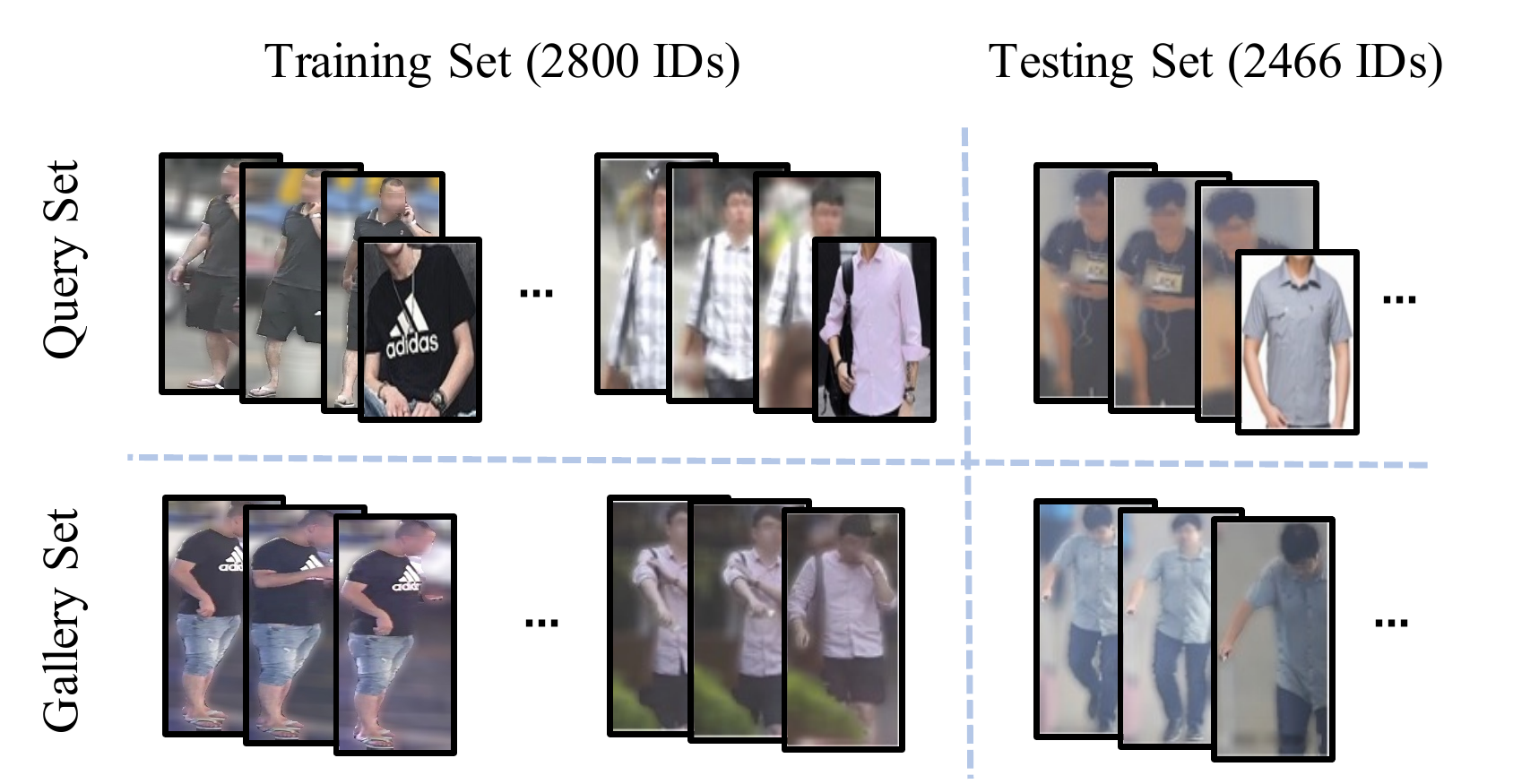}
    \caption{The partition of COCAS dataset. Vertical partition is to obtain training set and testing set according to person IDs. Horizontal partition divides COCAS into query set and gallery set according to clothes. Query set consists of query images and clothes templates while gallery set consists of target images.}
    \label{fig:split}
    \vspace{-0.2em}
\end{figure}

 \noindent \textbf{Person Association.} Now we have collected the data we need, but how to associate the images of the same person with different clothes is non-trivial. It is awkward to annotate images one by one from such an enormous database of images. 
 As shown in Fig. \ref{fig:process}, the association has 4 main steps: \emph{Person Clustering}: we cluster the similar person images based on the re-id feature, and manually remove the outliers images of different persons in the cluster. \emph{Face Detection} \cite{zhang_detect_face}: we select one image as an anchor image from each cluster and detect the face images from the anchor images. \emph{Face Retrieval}: we extract the facial feature by FaceNet \cite{Schroff_2015_CVPR_Face} and search the top-k neighbouring anchor images for each anchor image. \emph{Manually Annotation}: we visualize the body images corresponding to the anchor images, and manually select the truly matched neighbouring images. Based on the association results, our dataset is arranged as follows. For each person, we select 2 or 3 different clothes where each type of clothes has 2$\thicksim$5 images. Images of one kind of clothes are moved to the gallery set as the target images while other kinds are moved to query set as the query images. The partition as horizontal partition is illustrated in Fig. \ref{fig:split}.

 

 \noindent \textbf{Privacy Protection.}  We blur the specific regions of the selected body images to protect the personal information, including the faces, time and locations.  In greater details, MTCNN \cite{zhang_detect_face} has been used to get the bounding box of faces, and LIP \cite{Gong_2017_CVPR} is also adopted to separate the background and body regions. We then apply the gaussian filter to blur both facial and background regions, and we call the blurred version desensitized COCAS. 
  The experiments (section \ref{AblationStudies}) show that the performance will drop a little if we use desensitized COCAS, but we believe the desensitized COCAS is still valuable. This is because the faces cannot be always clear and background should not be a discriminative factor for the realistic re-id problem. In this paper, most experiments are based on desensitized COCAS.

 \noindent \textbf{Variations.}
 We explain the variation of COCAS. Their statistics are plotted in Fig. \ref{fig:statistics}. (1) \emph{Indoor/Outdoor}. We divide all the person images into two sets, according to the places they are captured, including `indoor' (23\%) and `outdoor' (77\%). The indoor and outdoor indicates different illumination conditions. (2) \emph{Clothes/Person}. 2,264 identities (43\%) have 3 different clothes and 3,002 identities (57\%) have 2 different clothes.  (3) \emph{Occlusion}.  A person image with occlusion means that the image is occluded by some  obstacles like cars, trees or other persons.  We also regard the case that the region of person is outside the image as a kind of occlusion. The images are categorized to four sets, including `heavy occlusion' (6\%), `medium occlusion' (24\%), `slight occlusion' (18\%) and `no occlusion' (52\%).

 \noindent \textbf{Protocols.}   Experimental protocols are defined as follows. Images of 2,800 persons are used for training, and the images of the remaining 2,466 persons are used for testing, which can be seen in Fig. \ref{fig:split}. In testing, we take 15,985 images selected from the 2,466 persons as the query images, and take the other 12,378 images as the target images forming the gallery of testing set. We search the target images with both the query images and the clothes templates.  Since a query image has multiple target images in the gallery set and CMC (Cumulative Matching Characteristic) curve can only reflect the retrieval precision of most similar target images. We additionally adopt mAP (mean Average Precision) that can reflect the overall ranking performance w.r.t. all the target images. 

\begin{figure*}[t]
   \centering
   \includegraphics[width=1.0\textwidth]{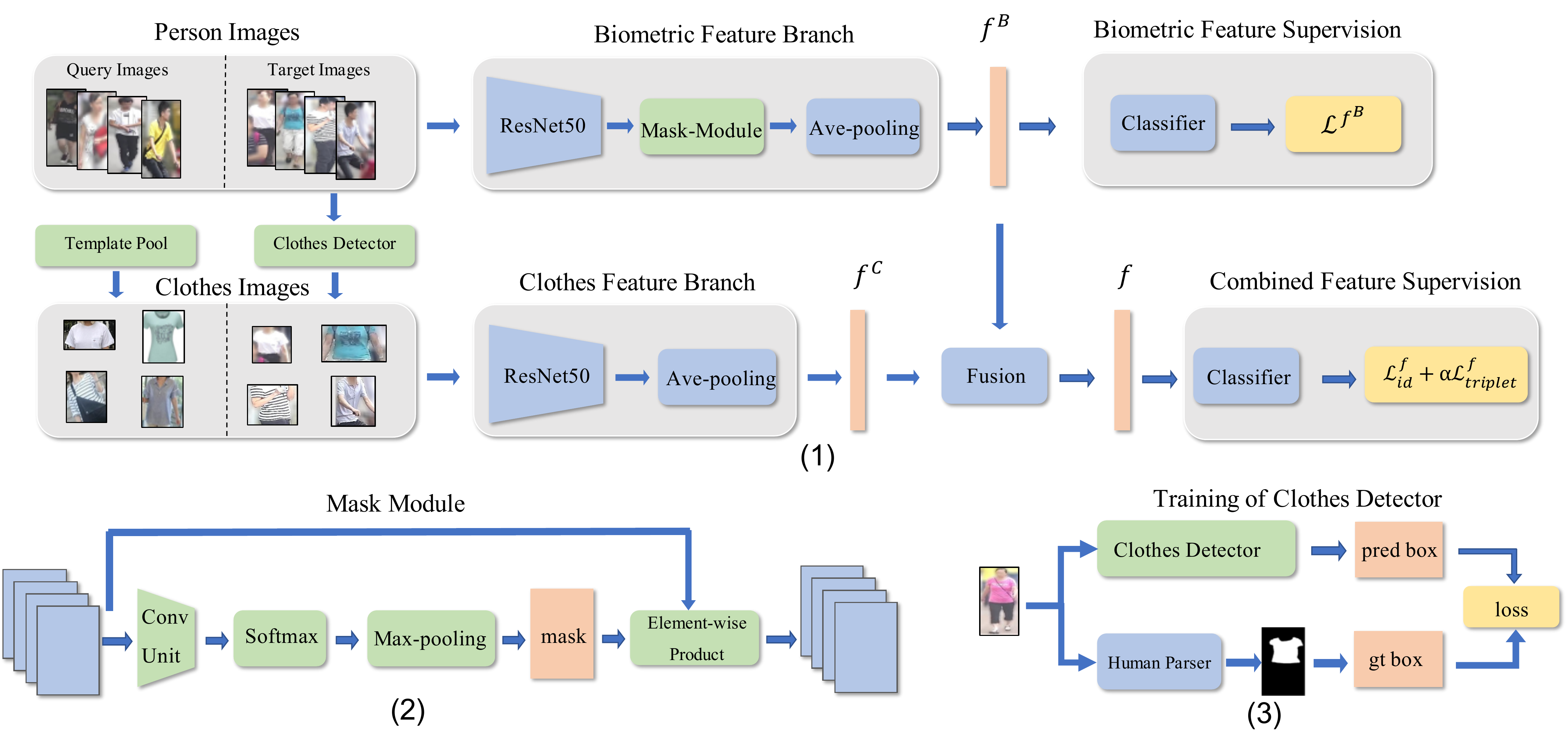}
   \caption{(1) The architecture of BC-Net. It contains two main branches including biometric feature branch and clothes feature branch. At the ends of these two streams, biometric feature and clothes feature are concatenated and then passed through a fully-connected layer to obtain the final feature with 256 dimensions. Note that the clothes detector based on faster RCNN is used to obtain clothes patches from target images. (2) The details of mask module. After convolution layers, the feature maps are normalized by softmax operation with each channels, then channel-wise max-pooling is applied to obtain the mask. At last,  biometric feature is selected by an element-wise product between the mask and the input feature maps. (3) The training process of the clothes detector. LIP, a human parsing model, is applied to obtain the clothes bounding boxes of person images rather than manual annotation.}
   \label{fig:network}
   \vspace{-0.5em}
\end{figure*}

\section{Methodology} \label{Baseline Approach}

According to our protocol, we need to search the target image from the gallery set by a similar clothes template and the person's another image with different clothes. Intuitively, the biometric traits in the query image and the appearance of the clothes template are helpful to search the target image. Therefore, we propose the two branch Biometric-Clothes Network (BC-Net) : one branch extracts the person biometric feature, and the other extracts the clothes feature. The biometric feature branch takes the person image as the input, and employs a mask module to better exploit the clothes irrelevant information. The clothes feature branch takes the clothes image, either the clothes template or the detected clothes patch, as the input, to generate the clothes feature. The final person representation combines the biometric feature and the clothes feature.

\subsection{Network Structure}
BC-Net has two branches, aiming to extract biometric feature  and clothes feature, respectively. The holistic architecture of BC-Net can be seen in Fig. \ref{fig:network}.

\noindent \textbf{Biometric Feature (BF) Branch}.  BF module takes a person image $I^{p}$ as input and employs ResNet50 \cite{He_2016_CVPR} as the backbone to yield feature maps $\mathbf{A}^p\in\mathbb{R}^{ H \times W \times D}$, where $H, W$ are the size of the feature map and $D$ is the feature dimension. To better exploit clothes irrelevant feature from more specific regions of the person, we further design a mask module as demonstrated in Fig. \ref{fig:network}. The module intends to emphasize the biometric feature while suppressing the feature of clothes and background. To obtain the mask $\mathbf{M}^p\in\mathbb{R}^{H \times W \times 1}$, $\mathbf{A}^p$ is first reduced to n-channel feature maps by three $1\times 1$ convolution layers, and then each feature map is normalized by a softmax function, which takes all the $H \times W$ values as input vector. Max-pooling along channels is applied to reduce n-channel feature maps to 1-channel feature map, yielding the mask. Based on the mask $\mathbf{M}^p$, the biometric feature ${f}^{B}\in\mathbb{R}^{D}$ is the obtained by the average pooling of the filtered feature map: 
\begin{equation}
    {f}_{k}^{B}  =  \frac{1}{H \times W}\sum_{i=1}^{H} \sum_{j=1}^{W} [\mathbf{A}^p_k \circ \mathbf{M}^p ]_{i,j}, 
\end{equation}
where $\circ$ indicates the element-wise product, $\mathbf{A}^p_k \circ \mathbf{M}^p$ is the \textit{k}th channel map of filtered feature maps and $f^{B}_k$ is the \textit{k}th element of $f^{B}$.

\noindent \textbf{Clothes Feature (CF) Branch}. CF branch tries to extract clothes related information. As our setting doesn't provide the clothes template for the target image, and we would like to process both query images and target images with the same network,  thus a clothes detector is additionally employed for the target image. The clothes detector is based on Faster RCNN \cite{ren2015faster}, which predicts the bounding box of the clothes from the target images. Either clothes template images or detected images are resized to the same size, and are fed into the CF branch as the input clothes image $I^{c}$. The CF branch also takes the ResNet50 as the backbone architecture, and outputs the clothes feature ${f}^{C}\in\mathbb{R}^{D}$ by average pooling over the feature maps $\mathbf{A}^c$.

The biometric feature ${f}^{B}$ and corresponding clothes feature ${f}^{C}$ are concatenated, then we estimate the feature vector ${f} \in \mathbb{R}^{d}$ by a linear projection:
\begin{equation}
 {f} = \mathbf{W}[({f}^B)^{\top}, ({f}^C)^{\top}]^{\top} + \mathbf{b},
\end{equation}
where $\mathbf{W}\in\mathbb{R}^{d\times 2D}$ and $\mathbf{b}\in \mathbb{R}^{d}$. ${f}$ is finally normalized by its $L_{2}$ norm in both training and testing stages.

\subsection{Loss Function}

We employ both identification loss and triplet loss over the training samples. The $n$th training sample is indicated by  $\mathcal{I}_{n} = \{I^{p}_{n}, I^{c}_{n}\}$, which consists of a person image and a clothes image. For a query person image, the clothes image is the clothes template describing the clothes in the target image. While for a target person image, the clothes image is the clothes image patch detected from itself. 

The combined feature $f_n$ can be regarded as a feature describing the target image, thus the conventional identification loss is employed for the combined features. Let $\mathcal{D}= \{ \mathcal{I}_{n}\}_{n=1}^{N}$ indicate the training samples, we make use of the ID information to supervise the combined feature:
\begin{equation}
\mathcal{L}_{id}^{f} = -\frac{1}{N}\sum_{n=1}^{N}\sum_{l=1}^{L} y_{n,l} \log \left( \frac{ e^{\bold{w}_{l}^{\top}f_n}}{ \sum_{m=1}^{L} e^{\bold{w}_{m}^{\top}f_n}} \right),
\end{equation}
where $\mathcal{D}$ has $N$ images belonging to $L$ persons. If the $n$th image belongs to the $l$th person, $y_{n,l}=1$, otherwise $y_{n,l}=0$. The parameters $\mathbf{w}_{l}$  associated with the feature embedding of the $l$th person. 

We now define a distance function $d(\mathcal{I}_{i}, \mathcal{I}_{j}) = \|f_i\!-\!f_j\|_{2}^{2}$, and further employ triplet loss to optimize inter-sample relationships. Let the $i$th triplet be $ \mathcal{T}_{i} = 
(\mathcal{I}^{a}_{i}, \mathcal{I}^{b}_{i}, \mathcal{I}^{c}_{i})$, and $\mathcal{I}^{a}_{i}$ is an anchor sample. $\mathcal{I}^{b}_{i}$ and $\mathcal{I}^{a}_{i}$ belong to same class while $\mathcal{I}^{c}_{i}$ and $\mathcal{I}^{a}_{i}$ are from different identities.  The triplet loss is defined as:
\begin{equation}
    \mathcal{L}_{triplet}^{f} = \frac{1}{N_{triplet}}\sum_{i=1}^{N_{triplet}} [d(\mathcal{I}^{a}_{i},  \mathcal{I}^{b}_{i}) \!+\! \eta \!-\!d(\mathcal{I}^{a}_{i},  \mathcal{I}^{c}_{i})  ]_{+},   \label{loss:triplet}
\end{equation}
where $N_{triplet}$ is the numbers the distance of positive pair to be smaller than the distance of negative pair at least by a margin $\eta$. The overall loss $\mathcal{L}^{f}$on the combined feature $f$ is the sum of the $\mathcal{L}_{id}^{f}$ and $\mathcal{L}_{triplet}^{f}$ defined as follows:
\begin{equation}
    \mathcal{L}^{f} = \mathcal{L}^{f}_{id} + \alpha \mathcal{L}^{f}_{triplet}. \label{loss:combine}
\end{equation}
To better learn the biometric feature, we additionally impose an identification loss, denoted by $\mathcal{L}^{f^{B}}$.

\subsection{Network Training} \label{Network Training}
 In BC-Net, the clothes detector and feature extractor are trained separately. 
 
 \noindent \textbf{Clothes Detector Training.}   The clothes detector is based on Faster RCNN \cite{ren2015faster}.  
 Instead of annotating the clothes bounding boxes manually,  we employed LIP \cite{Gong_2017_CVPR}, an effective human parsing model. For each image in the training set, We utilize LIP to produce the clothes mask, then calculate two coordinates of the left-up corner and right-bottom corner as the ground truth bounding box.  Stochastic gradient descent (SGD) is applied with momentum 0.9 for 30 epochs. 4 GPUs are employed for detector training and each GPU is set with a batch size of 12.

 \noindent \textbf{Feature Extractor Training.}  
We employ SGD to optimize the feature extractor with a momentum of 0.9. The optimization lasts for 100 epochs, and the initial learning rate is 0.00035,  which is further decayed to 0.00005 after 40 epochs. 
4 GPUs are used for training, and the batch size of each GPU is set to 32, \emph{i.e.}, 32 person images and 32 corresponding clothes images.  The 32 samples are about 8 persons and each person has 4 samples. For triplet loss, we take each sample as anchor sample, and choose the farthest positive sample and the closet negative sample to compose a triplet.

\begin{table*}
 \begin{center}
  \begin{tabular}
    {c||c |c| c| c| c| c|c|c|c|c}

    \hline
    \multicolumn{1}{c||}{\multirow{2}{*}{\textbf{No.}}}&\multicolumn{6}{c|}{\textbf{Experiment Setup}}&\multicolumn{4}{c}{\textbf{Performance}}  \\   
                 \cline{2-11}
  &mask&feature&loss&clothes detector&dataset&metric&mAP&top-1&top-5&top-10 \\ \hline
     1.& \textit{w/} &\textit{BF+CF}&\textit{$\mathcal{L}^{f}+\mathcal{L}^{f^B}$}&\textit{faster RCNN}&\textit{desensitized}&\textit{Euclid}&46.8&49.3&64.0&71.4   \\ 
    \hline
    2.& \textit{w/} &\textit{BF+CF}&\textit{$\mathcal{L}^{f}+\mathcal{L}^{f^B}$}&\textit{faster RCNN}&\textit{desensitized}&\textit{Euclid+RR}&54.8&53.9&60.7&69.0   \\ 
    3.& \textit{w/} &\textit{BF+CF}&\textit{$\mathcal{L}^{f}+\mathcal{L}^{f^B}$}&\textit{faster RCNN}&\textit{desensitized}&\textit{XQDA}&57.2&59.4&74.7&81.8   \\ 
    4.&\textit{w/}&\textit{BF+CF}&\textit{$\mathcal{L}^{f}+\mathcal{L}^{f^B}$}&\textit{faster RCNN}&\textit{desensitized}&\textit{XQDA+RR}&68.5&66.3&72.9&79.9   \\ 
    \hline
    5.& \textit{w/} &\textit{BF+CF}&\textit{$\mathcal{L}^{f}+\mathcal{L}^{f^B}$}&\textit{faster RCNN}&\textit{raw}&\textit{Euclid}&52.8&55.3&69.5&76.1   \\ 
    6.& \textit{w/} &\textit{BF+CF}&\textit{$\mathcal{L}^{f}+\mathcal{L}^{f^B}$}&\textit{faster RCNN}&\textit{raw}&\textit{Euclid+RR}&63.7&62.3&68.0&76.2   \\ 
    7.& \textit{w/} &\textit{BF+CF}&\textit{$\mathcal{L}^{f}+\mathcal{L}^{f^B}$}&\textit{faster RCNN}&\textit{raw}&\textit{XQDA}&65.1&67.0&\textbf{80.0}&\textbf{85.7}   \\ 
    8.& \textit{w/} &\textit{BF+CF}&\textit{$\mathcal{L}^{f}+\mathcal{L}^{f^B}$}&\textit{faster RCNN}&\textit{raw}&\textit{XQDA+RR}&\textbf{75.4}&\textbf{73.3}&77.9&84.5   \\
    \hline
    9.& \textit{w/} &\textit{BF}&\textit{$\mathcal{L}^{f}+\mathcal{L}^{f^B}$}&\textit{faster RCNN}&\textit{desensitized}&\textit{Euclid}&12.2&12.4&20.2&25.2   \\ 
    10.& \textit{w/} &\textit{CF}&\textit{$\mathcal{L}^{f}+\mathcal{L}^{f^B}$}&\textit{faster RCNN}&\textit{desensitized}&\textit{Euclid}&28.7&27.6&45.0&55.3   \\ 
    \hline
    11.& \textit{w/} &\textit{BF+CF}&\textit{$\mathcal{L}^{f}$}&\textit{faster RCNN}&\textit{desensitized}&\textit{Euclid}&32.7&33.7&50.6&60.3  \\ 
    12.& \textit{w/} &\textit{BF+CF}&\textit{w/o triplet loss}&\textit{faster RCNN}&\textit{desensitized}&\textit{Euclid}&42.8&44.8&59.2&66.5   \\ 
    \hline
    13.& \textit{w/o} &\textit{BF+CF}&\textit{$\mathcal{L}^{f}+\mathcal{L}^{f^B}$}&\textit{faster RCNN}&\textit{desensitized}&\textit{Euclid}&43.6&45.8&60.6&67.9   \\ 
    \hline
    14.&\textit{w/} &\textit{BF+CF}&\textit{$\mathcal{L}^{f}+\mathcal{L}^{f^B}$}&\textit{None}&\textit{desensitized}&\textit{Euclid}&39.5&41.0&55.7&63.4   \\ 
  \hline
 \end{tabular}
 \vspace{0.5em}
 \caption{Evaluation of our method on the COCAS dataset. We study the influence of mask, different features, loss function, clothes detector, desensitization, and different similarity metrics. Top-1, 5, 10 accuracies and mAP(\%) are reported. \emph{BF} and \emph{CF} denote the biometric feature and the clothes feature respectively. The combined feature is denoted by \emph{BF+CF}.}
 \label{Tab:Ablation_study}\vspace{-0.5em}
 \end{center} 
\end{table*}

\begin{figure*}[t]
\centering
    \begin{minipage}{0.49\textwidth}
    \centering
        \includegraphics[width=1.0\textwidth]{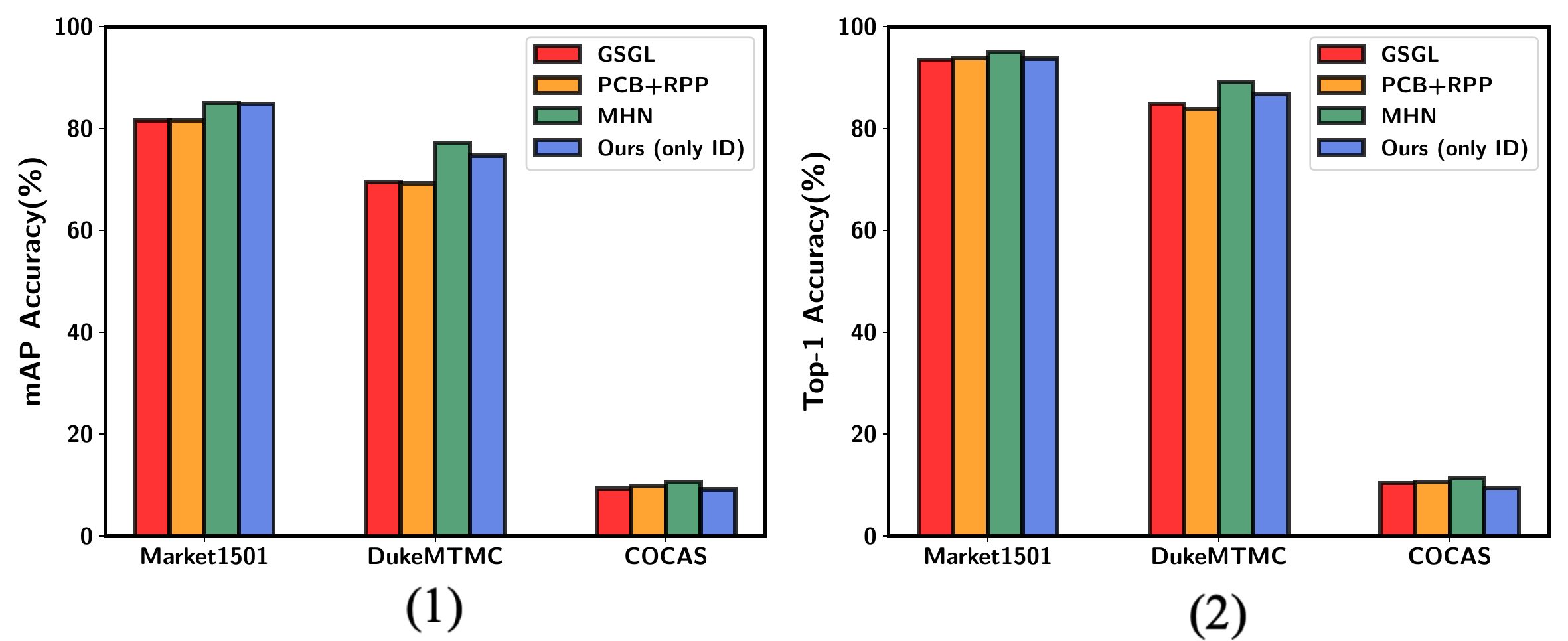}
        \figcaption{Training different datasets with SOTA methods and ours. (1) and (2) show the results of different methods on different methods in terms of
        the mAP and top-1 accuracy. Involving SOTA methods are GSGL \cite{Chen_2018_CVPR}, PCB+RPP \cite{Sun_2018_ECCV} and MHN \cite{Chen_2019_ICCV}.}
        \label{fig:onlyID}
    \end{minipage}
\hspace{0.5em}
\begin{minipage}{0.49\textwidth}  
    \centering
    \includegraphics[width=1.0\textwidth]{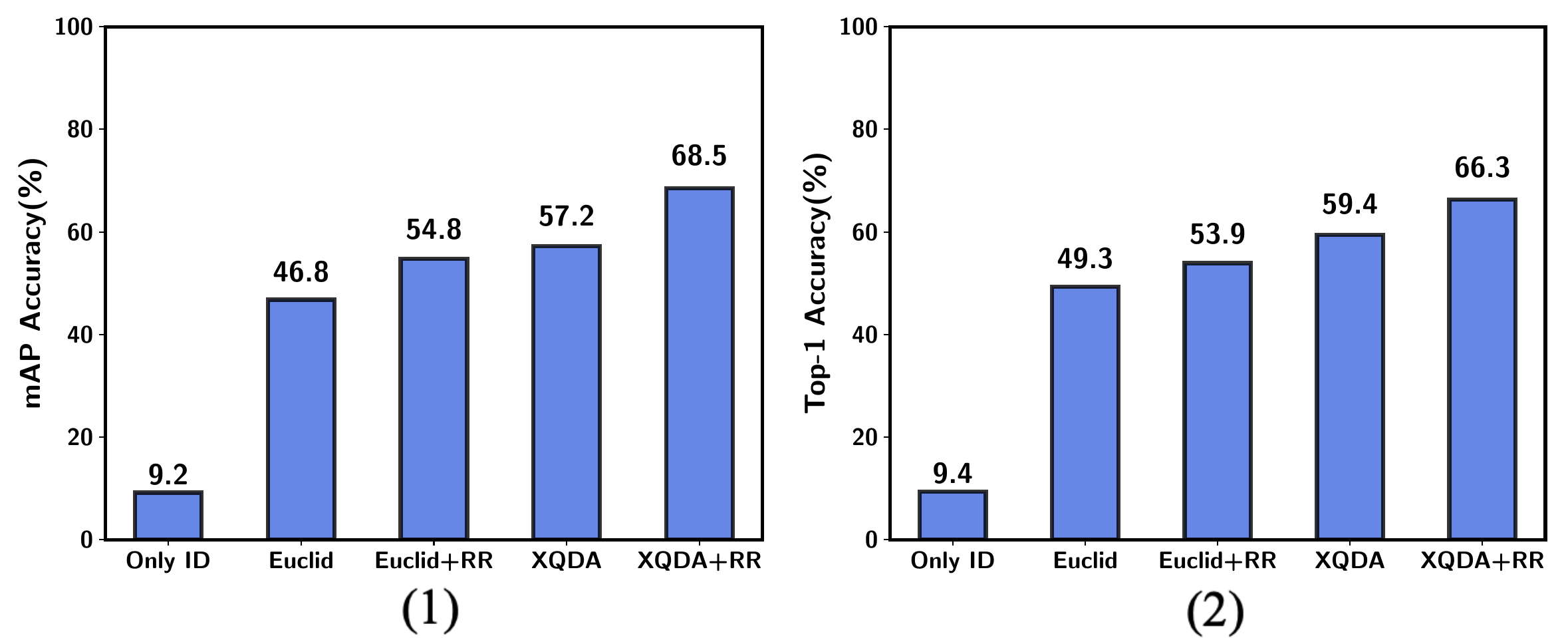}
     \figcaption{Training with provided clothes templates. A significant gap between using only ID and using provided clothes templates is shown above. We also demonstrate the effectiveness of XQDA\cite{liao2015person} and re-ranking\cite{zhong2017re}.}
    \label{fig:templates}
\end{minipage}

\end{figure*}


 \begin{figure*}[t]
    \centering
    \includegraphics[width=1.0\textwidth]{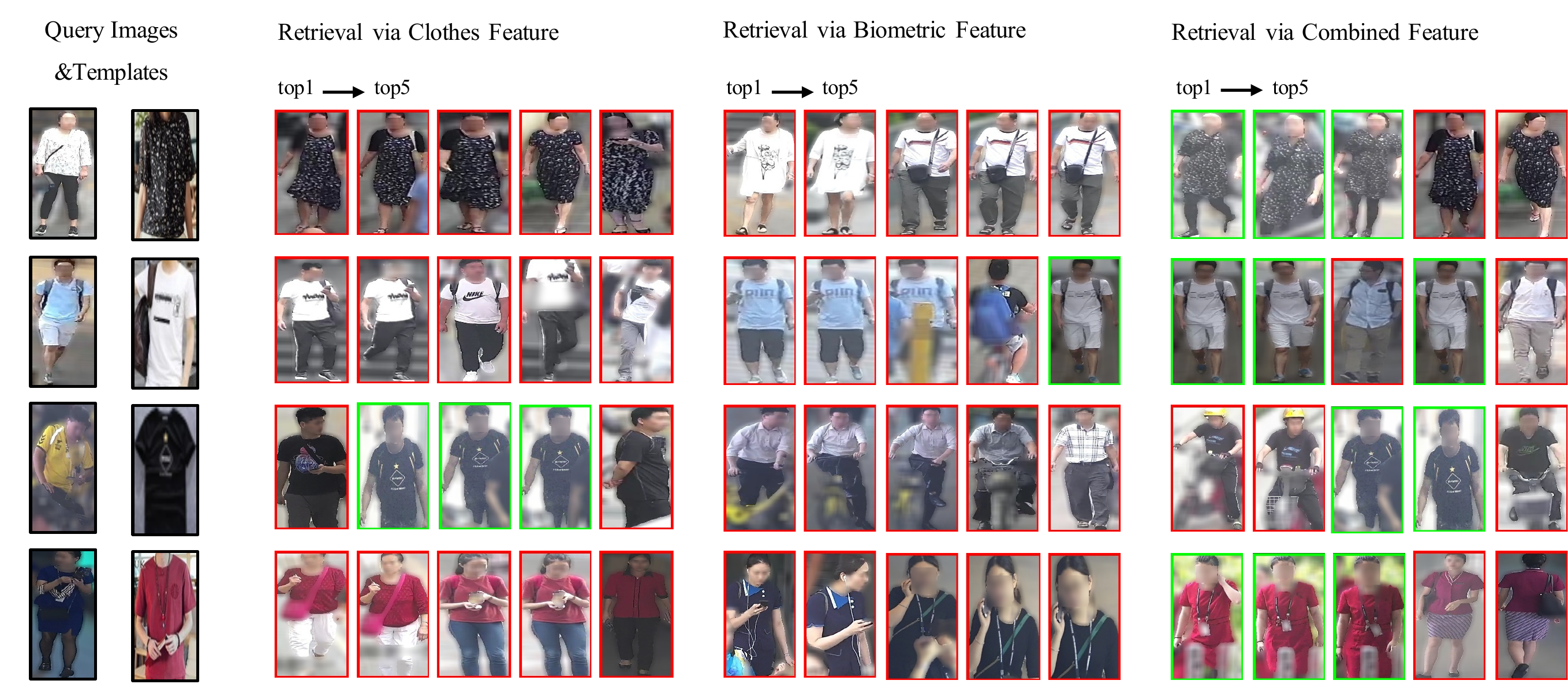}
    \caption{Examples of retrieval results by clothes feature, biometric feature and the combined feature, respectively. Green box means the images belong to the same identity with the query images, while the red box indicates the incorrect retrieved
    results. As shown above, the images retrieved via clothes feature have similar clothes with templates, \textit{e.g.}, red clothes find red clothes. While if only using biometric feature, the images that have similar clothes or body shape will be found. Combined feature is effective to find the images which have both characteristics of query images and clothes templates.} 
    \label{fig:retrieval}
\end{figure*}
\begin{figure*}[t]
    \centering
    \includegraphics[width=1.0\linewidth]{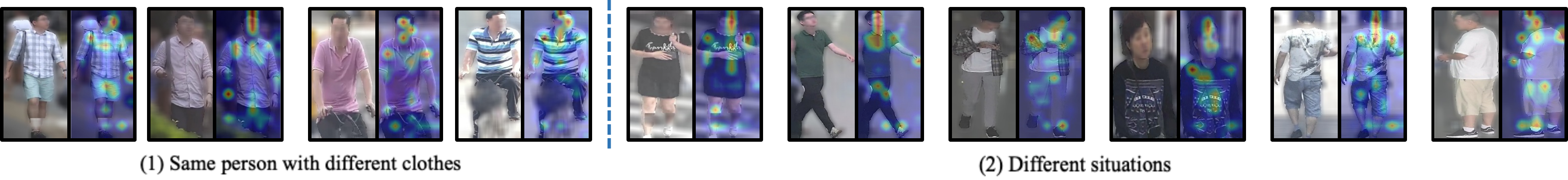}
    \caption{Visualization of masks. (1) shows the masks of the same person with different clothes. (2) shows the masks which are generated from images with different situations, including front, sideways, looking down, occlusion, back, \textit{etc.}}
    \label{fig:heatmap}
\end{figure*}

\section{Experiments} \label{Experiments}

In the experiment,  we first apply the current state-of-the-art approaches to the COCAS as well as other person re-id datasets only considering the identity information. Then we demonstrate how our method can improve the performance on COCAS by employing the clothes template and other post-processing strategies. Extensive ablation studies are conducted to evaluate the effectiveness of different components in our method.

\noindent \textbf{Implementation Details.}  The input person images are resized to $256 \times 128$ and the input clothes templates are resized to $128 \times 128$.  Random cropping, flipping and erasing are used for data augmentation. The margin $\eta$ in Eq. \ref{loss:triplet} is set to 0.3, The loss balance weights of $\alpha$ is set to 1.0.

\subsection{Overall Results}

\noindent \textbf{Learning with only ID labels.} First, we treat COCAS as a common person re-id dataset with only ID labels, \emph{i.e.}, without clothes templates. To highlight the dataset difference, we also incorporate Market1501\cite{zheng2015scalable} and DukeMTMC\cite{Zheng_2017_ICCV_Duke} for comparison.  All the datasets follow the standard training and testing partition protocol. Without employing additional clothes templates, our method treats all the images equally by detecting the clothes image patch from the original images and feeding them to the clothes feature branch. 
The results of several state-of-the-art (SOTA) and ours are shown in Fig. \ref{fig:onlyID}. It can be seen that our method can perform equally well with SOTAs on existing datasets, and all the methods obtain inferior results without utilizing the clothes templates.

\noindent \textbf{Learning with provided clothes templates.} We now involve the clothes templates for training. In particular, the query image takes the provided the template for the clothes branch and the target image utilizes the detected clothes patch. After training, we obtain the combined feature ${f}$, which is further normalized by its $L_{2}$ norm. Compared with the feature only trained with ID labels, the combined feature significantly improves the results even the similarity is measured by the Euclidean distance. As shown in Fig. \ref{fig:templates}, it achieves 37.6\% and 39.9\%  mAP and top-1 gains, respectively. We further study the effectiveness of two different similarity measuring schemes, \emph{i.e.}, the metric learning method (XQDA) \cite{liao2015person} and the re-ranking method (RR) \cite{zhong2017re}.  Results in Tab. \ref{Tab:Ablation_study}-2,3,4 show that XQDA and RR are effective and complementary. XQDA and RR improve the Euclidean feature distance by 8\% and 10.4\% mAP,  and their combination achieves 21.7\% mAP gains.

\subsection{Ablation Study}\label{AblationStudies}

In this section, we try to figure out what information is crucial to the clothes changing re-id. We also investigate various factors that can significantly influence the accuracy, including loss functions and clothes detector. 

\subsubsection{Performance Analysis}

\noindent \textbf{Biometric Feature v.s. Clothes Feature.}
To evaluate the effectiveness of the biometric feature and the clothes feature, we construct two variants for comparison. One only utilizes the biometric feature, and sets the clothes feature before fusion to be zero.  The other utilizes the clothes feature in a similar manner. As shown in Tab. \ref{Tab:Ablation_study}-9,10, only employing biometric feature or clothes feature leads to inferior results, whose mAP drops 34.6\% and 18.1\%, respectively. Note that the results of clothes feature are better than biometric feature, which indicates the clothes appearance is more important. Besides, the biometric feature is indispensable and complementary to the clothes feature, the final performance significantly boosts when combining the two features together. Fig. \ref{fig:retrieval} demonstrates several retrieval results generated by the three features. It can be seen that the biometric feature is independent with the clothes appearance and the combined feature can actually achieve better performance.

\noindent\textbf{Mask Module.}
To better obtain the biometric feature, the mask module is employed in the  biometric feature branch. Quantitatively, the mask module improves mAP from 43.6\% to 46.8\% and top-1 from 45.8\% to 49.3\% in Tab. \ref{Tab:Ablation_study}-1,13.  We also visualize the mask over the original image in Fig. \ref{fig:heatmap}, which indicates the mask mainly focuses on the facial and the joint regions. Although the facial region is desensitized in COCAS, it still serves as an important biometric clue. Meanwhile, the joint regions are potentially related to the pose or the body shape of a person.  

\noindent \textbf{Influence of Desensitization.}  In COCAS, we have obscured the faces and backgrounds of all images for privacy protection. As the facial region conveys important biometric information, we also train BC-Net with raw COCAS. The results can be seen in Tab. \ref{Tab:Ablation_study}-5,6,7,8. Compared with the results of desensitized COCAS in Tab. \ref{Tab:Ablation_study}-1,2,3,4, it improves about 6\% $\thicksim$ 9\% mAP when using the same similarity metric, indicating the desensitization actually weakens the biometric information. Nevertheless, the facial information is still helpful as has been analyzed in the mask module.

\subsubsection {Design Choices and Alternatives}

\noindent\textbf{Loss Function.}
As described in sec \ref{Network Training}, BC-Net is trained by the loss functions over both biometric feature and the combined feature.
We construct two variants.  The first removes the loss imposed over the biometric feature, \emph{i.e.}, training network only with $\mathcal{L}^{f}$. The second removes the triplet loss term in $\mathcal{L}^{f}$, \emph{i.e.}, training network with the loss of  $\mathcal{L}_{id}^{f} + \mathcal{L}^{f^B}$. The results are reported in Tab. \ref{Tab:Ablation_study}-11,12. Without $\mathcal{L}_{triplet}^{f}$, the performance decreases 4.0\% mAP and 4.5\% top-1. While without $\mathcal{L}^{f^B}$, the mAP drops sharply from 46.8\%  to 32.7\% and the top-1 accuracy drops from 49.3\% to 33.7\%. The results show that $\mathcal{L}^{f^B}$ is crucial to better extract the fine-grain biometric feature and filter irrelevant features out.  

\noindent\textbf{Clothes Detector.}
In BC-Net, we should first train the clothes detector, then use it to train the holistic network. To evaluate whether the clothes detector is necessary, we simply remove the clothes detector. If the person images are target images, the person images will be directly fed into both BF branch and CF branch.  As the results shown in Tab. \ref{Tab:Ablation_study}-14, without clothes detector, our method achieves 39.5\% of mAP and 41.0\% top-1, which drops 7.3\% and 8.3\% respectively. The clothes detector potentially removes the influence of other regions, such as the background or the trousers.


\section{Conclusion}
We have introduced a new person re-id benchmark considering the clothes changing problem, where each query is composed of a person image and a clothes template image. The benchmark contains over 60k images from 5,266 persons, where each identity has multiple kinds of clothes. For this re-id setting, we proposed the Biometric-Clothes Network, which can extract the biometric feature and the clothes feature, separately. Experiments have shown that traditional re-id methods perform badly when meeting clothes changing. While our method works well by utilizing the clothes templates.  The proposed setting and solution is promising in tracking suspects and finding lost children/elders in real-world scenarios.

{\small
\bibliographystyle{ieee_fullname}
\bibliography{egbib}
}

\end{document}